\documentclass[conference]{IEEEtran}
\usepackage{graphicx,
	psfrag,
	epsfig,
	epstopdf,
    enumerate,
	amsthm,
    amsmath,
	amssymb,
	url,
	algorithm,
	algorithmic,
	balance,
	enumerate,
	color,
    url,
	setspace,
	tikz,
	pgfplots,
    geometry,
    placeins,
    cleveref,
    multirow,
    array,
}
\usepackage{amsmath}%[centertags][tbtags][sumlimits][nosumlimits][intlimits][namelimits][nonamelimits]
\usepackage[nospace,noadjust]{cite}
\newgeometry{left=1.5cm,right=1.5cm,bottom=1.5cm, top=1.6cm}

\newcommand{\ignore}[1]{}

\begin{document}
\title{DNS Typo-squatting Domain Detection: A Data Analytics \& Machine Learning Based Approach}
\author{
\IEEEauthorblockN{Abdallah Moubayed, MohammadNoor Injadat, Abdallah Shami, and Hanan Lutfiyya}
		
\IEEEauthorblockA{\IEEEauthorrefmark{1} Western University, London, Ontario, Canada \\
	e-mails: \{amoubaye, minjadat, abdallah.shami, hlutfiyy\}@uwo.ca
}\\
%\IEEEauthorblockA{\IEEEauthorrefmark{2} University of Sharjah, Sharjah, United Arab Emirates \\
%	e-mail: anassif@sharjah.ac.ae	
%%
%}
}
\maketitle

\begin{abstract}
	Domain Name System (DNS) is a crucial component of current IP-based networks as it is the standard mechanism for name to IP resolution. However, due to its lack of data integrity and origin authentication processes, it is vulnerable to a variety of attacks. One such attack is Typosquatting. Detecting this attack is particularly important as it can be a threat to corporate secrets and can be used to steal information or commit fraud. In this paper, a machine learning-based approach is proposed to tackle the typosquatting vulnerability. To that end, exploratory data analytics is first used to better understand the trends observed in eight domain name-based extracted features. Furthermore, a majority voting-based ensemble learning classifier built using five classification algorithms is proposed that can detect suspicious domains with high accuracy. Moreover, the observed trends are validated by studying the same features in an unlabeled dataset using K-means clustering algorithm and through applying the developed ensemble learning classifier. Results show that legitimate domains have a smaller domain name length and fewer unique characters. Moreover, the developed ensemble learning classifier performs better in terms of accuracy, precision, and F-score. Furthermore, it is shown that similar trends are observed when clustering is used. However, the number of domains identified as potentially suspicious is high. Hence, the ensemble learning classifier is applied with results showing that the number of domains identified as potentially suspicious is reduced by almost a factor of five while still maintaining the same trends in terms of features' statistics.
\end{abstract}


\section{Introduction}\label{Intro_dns}
\indent Domain Name System (DNS) protocol, is the standard mechanism for name to IP address resolution \cite{DNS_definition}. Name servers generally maintain complete name/address information about a particular zone in a file known as the zone file. Every zone needs to provide a primary and a secondary name server to enhance resiliency, redundancy and load balancing. Hence, the DNS protocol is one of the core components in today's and future Internet architecture given that it helps users locate servers and mailing hosts and directly impacts data \cite{DNS_definition,future_internet}.\\
\indent However, DNS suffers from lack of data integrity and origin authentication processes. This makes it vulnerable to a variety of security concerns and breaches as shown by the various DNS attacks in recent years \cite{DNS_vulnerabilities,DNS_vulnerabilities1}. For example, the distributed denial of service (DDoS) attack on Dyn in October 2016 was one of the largest attacks of this kind with a reported strength of 1.2 Terabytes/s (Tbps) \cite{DNS_attacks1,DNS_attacks2}. This attack managed to bring down a significant portion of America's Internet Service \cite{DNS_attacks1,DNS_attacks2}. Another example is the attack on a Brazilian Bank's website in which attackers redirected all of the traffic targeted to the bank's website to their own servers by changing the DNS registrations of all the bank's domains \cite{DNS_attacks3}.\\
\indent Among the many vulnerabilities and security challenges of DNS protocol is the issue of typosquatting. Typosquatting refers to the registration of a domain name that is extremely similar to that of an existing popular brand (ex: \textit{www.paypal.com} and \textit{www.paypa1.com}). The goal is to redirect unsuspecting users to malicious/suspicious websites by registering confusingly similar domain names that the user might not pay attention to \cite{DNS_vulnerabilities4}. This is particularly important given that it can be a threat to corporate secrets, information theft, or committing fraud \cite{DNS_vulnerabilities2}.\\% Moreover, it can be used to steal information or commit fraud \cite{DNS_vulnerabilities2}.\\
\indent For practical security and availability reasons it is important that DNS is able to tolerate failures and attacks \cite{DNS_vulnerabilities1}. That is why a variety of techniques have been proposed in the literature to combat and protect against failures and attacks. For example, the DNSSEC protocol, which is a set of security extensions added to the original DNS protocol, has been proposed to address some of the existing vulnerabilities by providing data integrity and origin authentication \cite{dnssec_protocol}.%\cite{dnssec_protocol,dnssec_protocol1,dnssec_protocol2}. 
However, DNSSEC is still prone to other attacks such as synchronization attacks and amplified denial of service attacks \cite{DNS_vulnerabilities1,dnssec_performance}. Therefore, it is crucial that new efficient detection algorithms are developed that can help identify malicious/suspicious queries and protect systems from the various attacks.\\
\indent In this paper, a machine learning-based approach is proposed to tackle the typosquatting DNS vulnerability. To that end, exploratory data analytics is first used to better understand the behavior and trends observed in eight features that characterize the domain name. Furthermore, a majority voting-based ensemble learning classifier that is based on five traditional supervised machine learning classification algorithms is proposed that can detect suspicious domains. Moreover, the observed trends are verified by studying the same features in an unlabeled dataset using unsupervised machine learning clustering algorithm and through applying the developed ensemble learning classifier.\\
\indent The remainder of this paper is organized as follows: Section \ref{challenges_dns} presents the main vulnerabilities and security challenges facing the DNS protocol. Section \ref{methodologies_dns} gives an overview about the methodologies that have been proposed in the literature as well as other potential methodologies to adopt. Section \ref{proposed_approach_dns} presents the proposed approach adopted in this work. Section \ref{dataset_description_dns} describes the two dataset used in this work including the extraction of the considered features. Section \ref{results_dns} discusses the experiment details and results. Finally, Section \ref{conc_dns} concludes the paper.
%\begin{figure*}[!tbp]
%	\centering
%	\includegraphics[scale=.48,trim=0cm 0.5cm 0cm 0cm]{Figures/DNS_vulnerabilities.jpg}
%	%\includegraphics[trim=0.5cm 11.5cm 0.5cm 0.5cm, clip,scale=.35]{Figures/DNS_vulnerabilities.pdf}
%	\caption{DNS Vulnerabilities and Challenges}
%	\label{vulnerabilities}
%\end{figure*}

\section{DNS Vulnerabilities \& Challenges}\label{challenges_dns}
\begin{figure}[!htbp]
	\centering
	\includegraphics[scale=.5,trim=0.5cm 0.5cm 0cm 0cm]{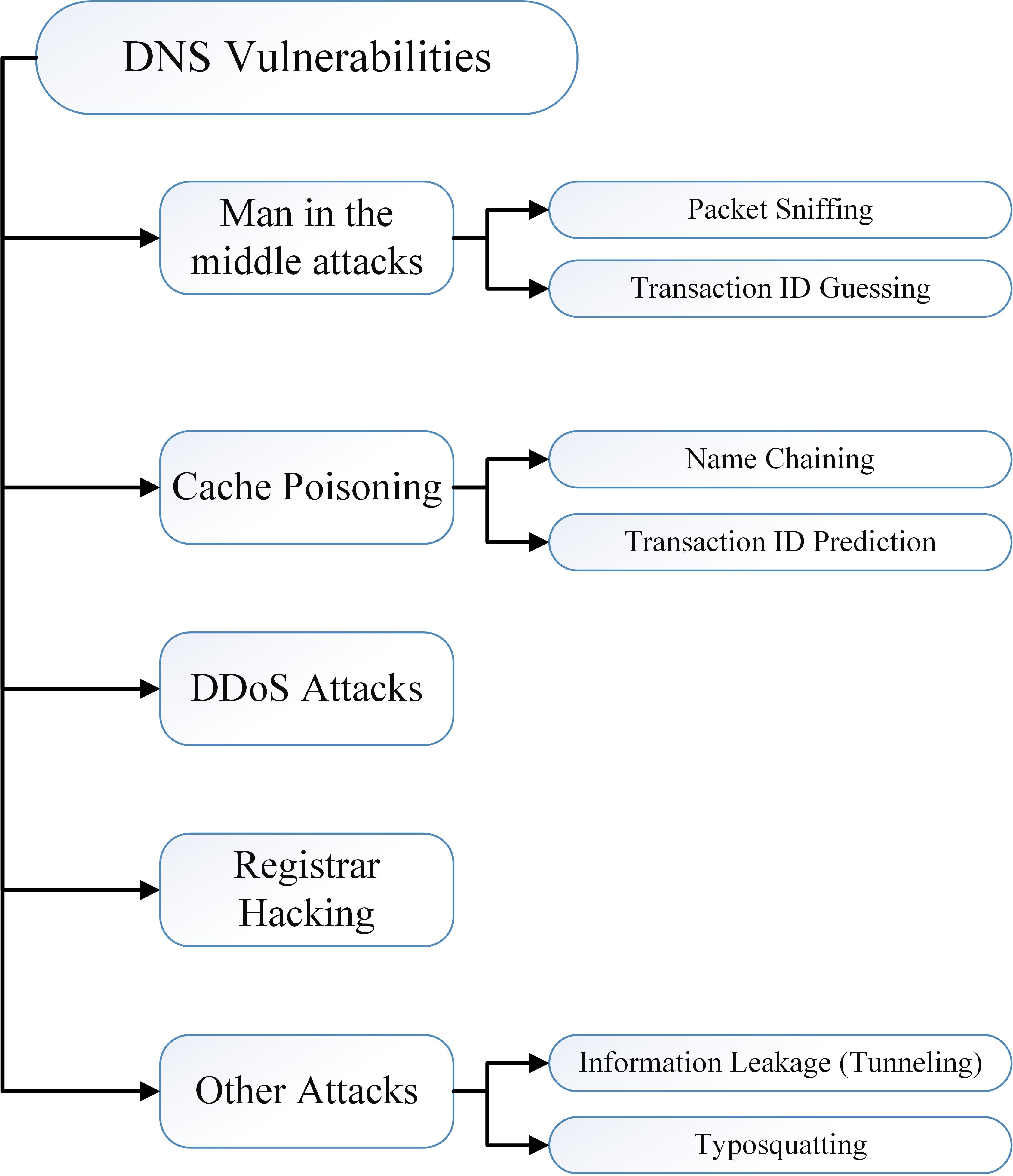}
	\caption{DNS Vulnerabilities and Challenges}
	\label{vulnerabilities}
\end{figure}

Due to its naivety, DNS protocol suffers from several vulnerabilities and security issues as it is prone to a variety of attacks. This is mainly due to the lack of data integrity and origin authentication processes within the  protocol. Fig. \ref{vulnerabilities} summarizes the most common vulnerabilities and attacks facing DNS protocol \cite{DNS_vulnerabilities1,DNS_vulnerabilities2,DNS_vulnerabilities3}.
\begin{enumerate}
\item Man in the middle attacks: DNS does not specify a mechanism for servers to provide authentication details for the data they push down to clients. Hence attacks such as Packet Sniffing or Transaction ID Guessing can occur \cite{DNS_vulnerabilities1}.
\begin{enumerate}[a-]
	\item Packet Sniffing: Attacker can capture DNS reply packets and modify them.
	\item Transaction ID Guessing: Guessing the transaction ID can allow the attacker to respond with false answers to legitimate queries.
\end{enumerate}
This is dangerous as it can threaten the users' privacy and direct them to suspicious domains and servers.
\item Cache Poisoning Problems: The current DNS protocol does not support any means to propagate data updates or invalidations to DNS caches in a fast and secure way. This could lead to Cache Poisoning by using Name Chaining or Transaction ID Prediction.
\begin{enumerate}[a-]
	\item Using Name Chaining: Attacker introduces false information into the DNS cache by adding arbitrary DNS names in the DNS response.
	\item Using Transaction ID Prediction: An attacker sends a large number of queries with domain names under his control. These replies have different transaction IDs. The attacker is hoping that one of the spoofed replies sent has the same transaction ID as that used between the two servers.
\end{enumerate}
\item DDoS attacks: Due to the hierarchal nature of the DNS architecture, root servers are prone to DDoS attacks that can cause a loss of availability of name resolution services. This in turn can lead to the stoppage of Internet service as illustrated by the DDoS attack on Dyn in October 2016 which was one of the largest attacks of this kind with a reported strength of 1.2 Terabytes/s (Tbps) that was able to bring down a significant portion of America's Internet service \cite{DNS_attacks1,DNS_attacks2}. 
\item Registrar hijacking: An attacker might hijack a registrar and gain control over all the domain names hosted by the registrar. This can potentially lead to losing the domain names of the enterprise/company. This was illustrated by the attack on a Brazilian Bank's website in which attackers redirected all of the traffic targeted to the bank's website to their own servers by changing the DNS registrations of all the bank's domains \cite{DNS_attacks3}. This attack affected thousands of users as their banking and even email and FTP credentials in some cases were obtained through this attack \cite{DNS_attacks3}. 
\item Other DNS attacks: other attacks exist such as, Information Leakage and DNS Dynamic Update.
\begin{enumerate}[a-]
	\item Information Leakage (DNS Tunneling): Using DNS queries or responses to leak information to a malicious user/server. Attacker could reveal sensitive information such as internal firewall configurations.
	\item Typosquatting: Registering a domain name that is extremely similar to that of an existing popular brand (ex: \textit{www.google.com} and \textit{www.goggle.com}). This can be used to steal information or commit fraud \cite{DNS_vulnerabilities2}.
\end{enumerate}
\end{enumerate}
This work focuses on one type of attack in particular, namely Typosquatting. As mentioned earlier, the danger of this attack is that it focuses on the users' attention as the change in a domain name might be minimal that they don't notice it. However, this change in domain name can have serious repercussions since it can lead to stealing of personal information, stealing corporate secrets, or committing fraud \cite{DNS_vulnerabilities2}. Hence, it is important to be have efficient and intelligent algorithms to detect such attacks.
\section{Previous \& Potential Methodologies}\label{methodologies_dns}
\subsection{Previous Methodologies}
\indent Internet security has always been a prime concern given the significant dependence on Internet services in our daily lives. With the growing number of attacks on Internet services, a need has risen to improve its security. Machine learning-based techniques have been proposed to help detect these attacks. For example, a variety of supervised machine learning classification algorithms such as support vector machines and artificial neural networks have been proposed in \cite{SDN_ddos_attack} to detect intrusion and DDoS attacks in Software-Defined Networks (SDN). Similarly, a decision tree-based classification algorithm is proposed in \cite{cloud_ddos_attack} to detect DDoS attacks in cloud environments.\\
\indent However, very few previous works consider DNS security using machine learning. One such work was proposed in \cite{DNS_vulnerabilities3} in which a decision tree-based classifier was also used to detect malicious/suspicious domains by looking at a variety of query level features such as query size and time to live (TTL). 
\subsection{Potential Methodologies}
\begin{figure}[!htbp]
	\centering
	\includegraphics[scale=.5,trim=0.5cm 0cm 0cm 0cm]{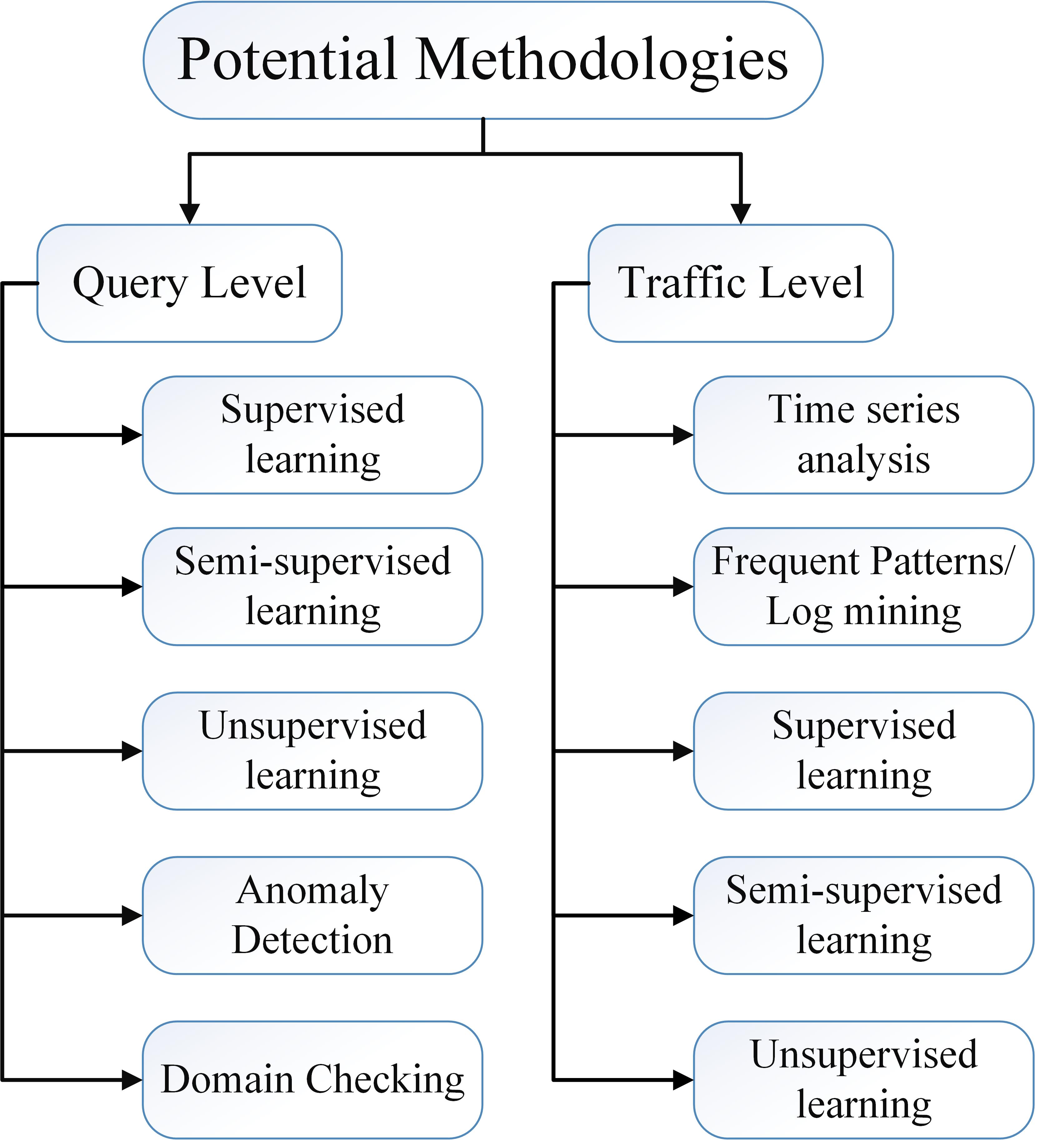}
	\caption{Possible Methodologies}
	\label{possible_methods}
\end{figure}

\indent There are various methodologies that can be used to detect malicious/suspicious DNS queries and domains. These methodologies can be divided into two main categories: methodologies adopted at the query level and methodologies adopted at the traffic level. For each level, a distinct set of features can be considered and studied. Fig. \ref{possible_methods} gives an overview of the different methodologies that can be adopted to tackle the issues facing DNS protocol and improve its security.
\subsection{Query Level:} 
\indent Each query is looked at individually and a decision is made using the adopted technique. A variety of techniques have been/can be adopted.
	\begin{enumerate}
	   \item \textbf{Supervised learning:} Supervised machine learning classification algorithms can be used on labeled datasets to train a classifier that is able to detect malicious/suspicious activity with high accuracy and low false positive rate. 
	   \item \textbf{Semi-supervised learning:} Semi-supervised machine learning algorithms can be used in cases where the dataset is a mixture of labeled and unlabeled data points. The knowledge gained from the labeled data points can be leveraged to predict the label for the unlabeled data points. 
	   \item \textbf{Unsupervised learning:} Unsupervised machine learning clustering algorithms can be used in cases where the labels are not available. In this case, data points that look similar (based on some distance metric) are group together. An expert with contextual knowledge would then make a decision on what each cluster means. 
	   \item \textbf{Anomaly Detection:} Anomaly detection can be used to identify malicious/suspicious activity by determining whether a data point is anomalous or not. Again, an expert with contextual knowledge is needed to define the normal behavior which will be used to determine the anomaly score of any new data point.
	   \item \textbf{Domain Checking:} The most basic way is to compare the domain with a list of malicious/suspicious domains such as the one available at \cite{malicious_domains_list}. However, this will depend on the update rate of this list since this methodology can become inefficient if the list becomes stale. This is further emphasized in \cite{domain_checking_inefficiency} as it was shown that this methodology is more reactive than proactive, which keeps attackers one step ahead. 
	\end{enumerate}
To adopt one of these methodologies, the following list gives some of the features that can be used to detect malicious/suspicious activity:
	\begin{itemize}
	\item Query size
	\item Number of IP addresses associated with a domain
	\item Type of domain requested 
	\item Length of domain name (number of characters in the domain name)
	\item Number of unique characters in domain name
	\item Number of numerical characters within domain name
	\item Percentage of numerical characters within domain name
	\item Percentage of unique characters from longest meaningful substring
	\item Age of domain (how old is the domain)
	\item Number of canonical names associated with a domain
	\item Time when domain is being accessed
	\item Registrar name
	\item Cost of registering domain with the registrar name
	\item Trustworthiness of registrar	
	\end{itemize}
%\vspace{length}
\subsection{Traffic Level:} 
\indent A group of queries are studied and analyzed using the adopted technique. A variety of techniques can be adopted.
	\begin{enumerate}
		\item \textbf{Supervised learning:} Supervised machine learning classification algorithms can also be applied at the traffic level. This can be done to detect specific types of attacks such as DDoS attacks. This is because such attacks can't be detected by one query due to their group-based nature. Features such as access ratio and number of distinct IP addresses associated with a domain can be used to build the classifier \cite{DNS_vulnerabilities3}. 
		\item \textbf{Semi-Supervised Learning:} Semi-supervised machine learning can also be used in this case by leveraging the information gained from a labeled dataset and applying it to an unlabeled dataset.
		\item \textbf{Unsupervised Learning:} Unsupervised machine learning clustering algorithms again can be used at the traffic level to group similarly looking data points. Contextual knowledge is needed as well to define what each cluster represents. 
		\item \textbf{Time Series Analysis:} Time series analysis can be used to determine any time-dependent correlations in the data. For example, it can be check if a specific pattern appears every day or every week. This can be hugely beneficial in trying to detect automated attacks that often follow a particular time-related pattern.
		\item \textbf{Frequent Patterns/Log Mining:} Frequent patterns and log mining techniques can also be used to identify any correlation between different frequent itemsets. Similar to the approach in \cite{frequent_patterns_mining}, association rules algorithms can be used to find statistical correlation between frequent itemsets. For example, log mining can be used to find if two domains that are being frequently queried are correlated in time.
	\end{enumerate}
To that end, the following list gives some of the features that can be used to detect anomalous/malicious activity:
\begin{itemize}
	\item Frequency of requests for different domains from one machine
	\item Frequency of requests to a domain from different machines
	\item Access ratio of a domain to the whole set of requests
	\item Time separation between consecutive domain requests from one machine
	\item Rate of change of IP address
	\item Average TTL value
	\item Standard deviation of TTL value
\end{itemize}     
\section{Proposed Approach}\label{proposed_approach_dns}
\subsection{Proposed Approach}
\indent This paper proposes a machine learning-based approach to tackle the typosquatting DNS vulnerability by detecting suspicious domain names. The proposed approach can be divided into three components as follows:
\begin{enumerate}
	\item Study and understand the characteristics of such domain names using exploratory data analytics techniques used on a labeled dataset (limited dataset).
	\item Develop a supervised machine learning-based classifier that can detect malicious/suspicious domains.
	\item Validate the observed trends by studying the same set of characteristics in a larger unlabeled dataset to ensure the extendability of the proposed approach. 
\end{enumerate}
\subsection{Application of the Proposed Approach}
\indent To understand the characteristics of suspicious domain names, eight features are extracted from the domain name that can be used to characterize it. The features where chosen specifically because of the typosquatting attack that is mainly directed towards modifying the domain name. Exploratory data analytics is then used to better understand the trends observed in these features. This is done by studying the statistics of these features, their probability density function, and their correlation with the class label (whether domain is legitimate or generated algorithmically).\\ 
\indent The second step is developing a majority voting-based ensemble learning classifier that is based on five traditional supervised machine learning classification algorithms. The considered algorithms are decision trees (C4.5), K-nearest neighbors (K-NN), logistic regression (LR), Naive Bayesian (NB), and Support Vector Machines (SVM). These algorithms were chosen due to their popularity and efficiency in various applications, especially in text-based classification. The goal is to improve the detection accuracy of the ensemble learning classifier by reducing the bias of the based classifiers and reduce the false positive detection of suspicious domains.\\ 
\indent The final step is verifying the trends observed in the labeled dataset. This is done by studying the same features in an unlabeled dataset using unsupervised machine learning clustering algorithm, namely the K-means algorithm. It is worth mentioning  that due to the size of the dataset, other clustering algorithms such as hierarchical clustering couldn't be implemented as the needed distance matrix would consist of more than 9 million values Moreover, the developed ensemble learning classifier is applied to the unlabeled dataset to predict the label of each domain.
\subsection{Complexity of Proposed Approach}
\indent The computational complexity of the proposed approach is governed by the complexity of each of its individual algorithms. To do so, it is assumed that the number of training samples is $M$ and number of features is $N$ with $M>>N$. The complexity of C4.5 algorithm is $O(M^2N)$ while the complexity of K-NN algorithm is $O(MN)$ \cite{complexity1,complexity2}. On the other hand, LR has a complexity of $O(N^3+MN^2)$ and NB algorithm's complexity is $O(MN)$ \cite{complexity1,complexity2}. Lastly, the SVM algorithm's complexity is $O(M^2N)$ while k-means algorithm has a complexity of $O(MNk)$ \cite{complexity1,complexity2}. Hence, the overall complexity of the proposed approach is $O(M^2N)$.  This is a polynomial running time complexity making it tractable and acceptable.
\subsection{Contributions}
\indent The contributions of this work can be summarized as follows:
\begin{itemize}
	\item The behavior and trends of several domain name-related features is studied using exploratory data analytics.
	\item A majority voting-based ensemble learning classifier based on five supervised machine learning classification algorithms is proposed to identify suspicious domain names that achieves higher accuracy.
	\item The observed trends are verified by studying the same set of features in an unlabeled dataset using unsupervised machine learning clustering algorithms and through applying the developed ensemble learning classifier.
\end{itemize}
\section{Dataset Description}\label{dataset_description_dns}
\indent This work considers two different datasets, a labeled dataset and an unlabeled dataset. In what follows, a brief description of both datasets is given.
\subsection{Data Preprocessing:}
\subsubsection{Labeled Dataset}\mbox{}\\
\indent The dataset was collected by the authors of the ``Data Driven Security'' book \cite{labeled_data_source1}. This was done by combining data collected from Alexa's top 1 million legitimate domains and from ``Cryptolocker'' for domains generated algorithmically (DGA) \cite{labeled_data_source}. The dataset consists of 133,926 unique domains of type A with 81,261 legitimate domains and 52,665 DGA domains. Each record consists of the following three fields:
\begin{itemize}
	\item Host: The complete url of the domain.\\ 
	(ex.: www.mydaily.co.uk)
	\item Domain: The domain accessed (ex.: mydaily).
	\item Domain Class: The classification of the domain (either legit or dga).
\end{itemize}
\subsubsection{Unlabeled Dataset}\mbox{}\\
\indent The unlabeled dataset used is based on the DNS Census 2013 public dataset \cite{unlabeled_data_source}. The dataset consists of 750,719,726 type A records (record having the original domain name registered and corresponding IPv4 address). However, only the first 1 million records are used in this work as the entire set of records couldn't be processed. Each record consists of the following two fields:
\begin{itemize}
	\item Domain: The domain accessed .
	\item IP4 address: IP4 address associated with the domain .
\end{itemize}
This dataset was then processed to extract 302,689 unique domains using MATLAB.
%\begin{table}[h]
%	\centering
%	\caption{Sample of Original Dataset}
%	\scalebox{0.85}{
%		\begin{tabular}{|p{2cm}|l|l|p{2cm}|p{2cm}|p{2.5cm}|}	\hline
%			Event Date &Event Type& Event Location&Session Start & Session End & Student ID \\ \hline
%			15-09-08 20:09:12&pres.begin&/presence/course-id-presence&15-09-08 20:09:03&15-09-09 02:32:43&student000000\\ \hline
%			15-09-08 20:09:49&lessonbuilder.read&/lessonbuilder/page/39397935&15-09-08 20:09:03&15-09-09 02:32:43&student000000\\ \hline
%			15-09-09 23:36:31&syllabus.read&/syllabus/course-id/1&15-09-09 23:34:59&15-09-10 02:58:00&student000001\\ \hline			
%	\end{tabular}}
%	\label{dataset_sample}
%\end{table}

%\vspace{-5mm}
\subsection{Data Transformation:}
\indent\indent Using MATLAB, both datasets where transformed from their original format into a new dataset consisting of eight domain name-based features. These features are used to characterize each unique domain. The features where chosen specifically because of the typosquatting attack that is mainly directed towards modifying the domain name. The extracted features are:
\begin{itemize}
	\item Length of Domain Name
	\item Number of Unique Characters 
	\item Number of Unique Letters
	\item Number of Unique Numbers
	\item Ratio of Letters to Domain Length	
	\item Ratio of Numbers to Domain Length
	\item Ratio of Unique Letters to Unique Characters
	\item Ratio of Unique Numbers to Unique Characters
\end{itemize}
In addition to these features, a binary feature is added to the labeled dataset to show the class of the domain. In this case, 1 was used to represent a DGA domain while 0 was used to represent a legitimate domain. It is worth noting that all these features are numeric with the first four being integers and the remaining being continuous. %Table \ref{table_of_metrics_dns} shows the value type and range of each of the aforementioned features.
%\begin{table}[h]
%	\centering
%	\caption{Domain Features Description}
%	\scalebox{0.85}{
%		\begin{tabular}{|p{4.5cm}|p{1.7cm}|p{1.5cm}|}	\hline
%			Feature & Value Type & Range of Values \\ \hline
%			Length of Domain Name&Numeric&[1,2,...,68]\\ \hline
%			Number of Unique Characters&Numeric&[1,2,...,36] \\ \hline
%			Number of Unique Letters&Numeric&[1,2,...,26]\\ \hline
%			Number of Unique Numbers&Numeric&[0,1,...,10]\\ \hline
%			Ratio of Letters to Domain Length&Numeric&[0-1]\\ \hline	
%			Ratio of Numbers to Domain Length&Numeric&[0-1]\\ \hline
%			Ratio of Unique Letters to Unique Characters&Numeric&[0-1]\\ \hline
%			Ratio of Unique Numbers to Unique Characters&Numeric&[0-1]\\ \hline
%			Domain Class (for labeled dataset)&Numeric&[0,1]\\ \hline
%		\end{tabular}
%	}
%	\label{table_of_metrics_dns}
%\end{table}
\section{Experiment Results \& Discussion}\label{results_dns}
\subsection{Experiment Setup}
%\indent\indent The following settings were used to conduct the experiments in this work:
%\subsubsection{Software}\mbox{}\\
%\indent\indent The following software was used to run the experiment and record the results:
%\begin{itemize}
%	\item Operating System: Microsoft Windows 10 (64-Bit OS, X-64 based processor)
%	\item MATLAB: MATLAB was used to transform the data from its original state to the new desired dataset representing the previously provided features.
%\end{itemize}
%\subsubsection{Experiment}\mbox{}\\
\indent\indent MATLAB was used to train the different classifiers considered in this work on a Microsoft Windows 10 (64-Bit OS, X-64 based processor) system, namely the C4.5 classifier, K-NN classifier, LR classifier, NB classifier, and SVM classifier. Moreover, the ensemble learning classifier is trained as a majority vote of these classifiers.
\subsection{Results \& Discussion}
\indent The experiment results are divided into two sections, one for the labeled dataset and one for the unlabeled dataset.
\subsubsection{{Labeled Dataset}}
\begin{figure}[!ht]
	\centering
	\includegraphics[scale=.45,trim=0.5cm 0cm 0cm 0cm]{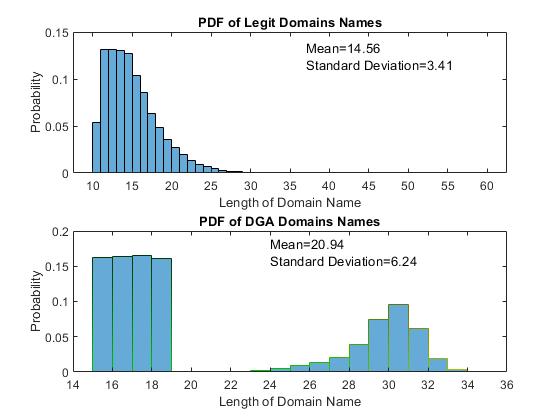}
	\caption{Probability Density Function of Domain Length For Labeled Dataset}
	\label{labeled_domain_length}
\end{figure}

\begin{figure}[!ht]
	\centering
	\includegraphics[scale=.45,trim=0.5cm 0cm 0cm 0cm]{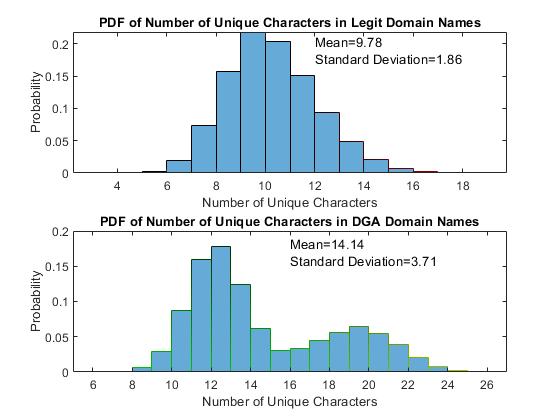}
	\caption{Probability Density Function of Number of Unique Characters For Labeled Dataset}
	\label{labeled_num_of_unique_characters}
\end{figure}

\paragraph{Exploratory Data Analytics}\mbox{}\\
\indent As mentioned earlier, exploratory data analytics is applied to the extracted features to better understand the behavior and trends observed for legitimate and DGA domains. Figs. \ref{labeled_domain_length} and \ref{labeled_num_of_unique_characters} show the probability density function of the domain name length and number of unique characters respectively for legitimate and DGA domains. It can be seen that legitimate domains tend to have shorter domain names and fewer number of unique characters. This is expected since legitimate domains need to be easily memorable. However, DGA domains tend to be more random which results in a higher number of unique characters and a wider distribution.
\mbox{}\\
\indent To further understand the behavior, the correlation between each feature and the domain class is studied. Table \ref{feauture_class_correlation} shows that the highest correlation with the domain class is that of the number of unique characters. This gain can be attributed to the fact that legitimate domains tend to have more memorable names with less characters while DGA domains tend to have more unique characters to increase the randomness of the domain name. 
\begin{table}[h]
	\centering
	\caption{Correlation Between Extracted Features and Domain Class}
	\scalebox{0.75}{
		\begin{tabular}{|p{4.5cm}|p{1.8cm}|}	\hline
			Feature & Correlation \\ \hline
			Number of Unique Characters & 0.663\\ \hline
			Number of Unique Letters& 0.653\\ \hline
			Length of Domain Name& 0.621\\ \hline
			Number of Unique Numbers& 0.329\\ \hline
			Ratio of Numbers to Domain Length& 0.281\\ \hline
			Ratio of Unique Letters to Unique Characters& 0.269\\ \hline
			Ratio of Unique Numbers to Unique Characters& 0.269\\ \hline
			Ratio of Letters to Domain Length& 0.242\\ \hline
		\end{tabular}
	}
	\label{feauture_class_correlation}
\end{table}
\paragraph{Ensemble Learning Classifier}\mbox{}\\
\indent An ensemble learning classifier was developed based on five traditional supervised machine learning classification algorithms using a majority-vote rule. This was done to improve the prediction and reduce the bias and variance of the underlaying classifiers \cite{ensemble_learner_benefits}. Four metrics are considered in evaluating the performance of the proposed ensemble learning classifier, namely the accuracy, precision, recall, and F-score using the same equations given in \cite{performance_metrics}. Table \ref{labeled_dataset_results} shows the results of the ensemble learning classifier as compared to that of its base classifiers.
\begin{table}[h]
	\centering
	\caption{Performance Evaluation of Classifiers}
	\scalebox{0.75}{
		\begin{tabular}{|p{1.8cm}|p{1.4cm}|p{1.4cm}|p{1.4cm}|p{1.1cm}|}	\hline
			Algorithm & Accuracy (\%) & Precision (\%)& Recall (\%)&F-score\\ \hline
			C4.5&88.1&84.5&\textbf{95.8}&0.89\\ \hline
			K-NN&88.2&83.8&94.3&0.89\\ \hline
			LR&79.9&83.8&77.3&0.80\\ \hline
			NB&88.1&84.5&\textbf{95.8}&0.89\\ \hline
			SVM&79.6&85.2&71.5&0.78\\ \hline
			Ensemble Learning Classifier&\textbf{88.4}&\textbf{85.5}&92.4&\textbf{0.89}\\ \hline
		\end{tabular}
	}
	\label{labeled_dataset_results}
\end{table}
The results show that the ensemble learning classifier has the highest accuracy, precision, and F-score while have a relatively high recall value. This shows that the proposed ensemble learning classifier is able to identify malicious/suspicious domains accurately due to the high precision rate. This emphasizes the efficiency of the proposed classifier when compared to other traditional classifiers.
\subsubsection{Unlabeled Dataset}
\begin{figure}[!t]
	\centering
	\includegraphics[scale=.45,trim=0.65cm 0cm 0cm 0cm]{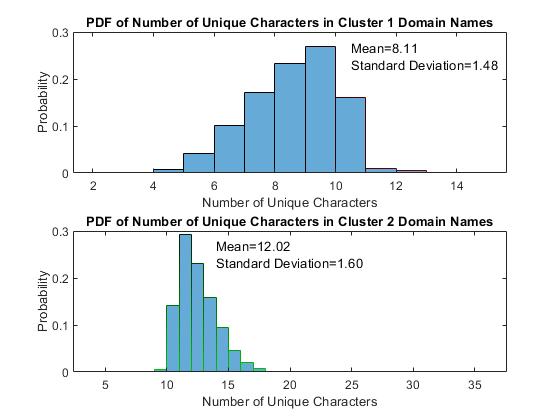}
	\caption{Probability Density Function of Number of Unique Characters For Unlabeled Dataset Using K-means Clustering}
	\label{unlabeled_num_of_unique_characters}
\end{figure}

\paragraph{Unsupervised Clustering}\mbox{}\\
\indent To further validate the observed trends, an unsupervised machine learning clustering algorithm is applied to an unlabeled dataset consisting of 302,689 unique domain entries. In particular, K-means algorithm was used to group the points into two distinct clusters. It is worth noting that due to the size of the dataset, other clustering algorithms such as hierarchical clustering couldn't be implemented as the needed distance matrix would consist of more than 9 million values.\\
\indent The clustering algorithm resulted in 165,593 points grouped into cluster 1 and 137,096 points grouped into cluster 2. Table \ref{2level_clustering_centroid} shows the centroid of each of the two clusters. It can be seen that cluster 1 on average has a smaller domain length and fewer number of unique characters (more specifically  fewer number of letters) while cluster 2 on average has longer domain names with a larger number of unique characters.
\begin{table}[!ht]
	\centering
	\caption{Centroid Means For Two Level Clustering Model}
	\scalebox{0.75}{
		\begin{tabular}{|p{5cm}|p{1.3cm}|p{1.3cm}|}	\hline
			%\backslashbox{Metric}{Level}& Low & High \\ \hline
			&\multicolumn{2}{c|}{Cluster}\\  \cline{2-3}
			Feature&Cluster 1&Cluster 2\\ \hline
			Domain length&12.3023&20.4739\\ \hline
			Num. of Unique Characters&8.1069&12.0255\\ \hline
			Num. of Unique Letters&7.9815&11.9744\\ \hline
			Num. of Unique Numbers&0.1253&0.0511\\ \hline
			Ratio of Letters to Domain Length&0.8741&0.9107\\ \hline
			Ratio of Numbers to Domain Length&0.0124&0.0026\\ \hline
			Ratio of Unique Letters to Unique Characters&0.9850&0.9964\\ \hline
			Ratio of Unique Numbers to Unique Characters&0.0150& 0.0036\\ \hline
		\end{tabular}
	}
	\label{2level_clustering_centroid}
\end{table}\\
\indent Moreover, the probability density function of the number of unique characters for each of the two clusters is plotted. Fig. \ref{unlabeled_num_of_unique_characters} shows that cluster 1 has a smaller mean and standard deviation when compared to that of cluster 2. Based on the statistics illustrated in this figure and the insights gained from the statistics of the labeled dataset, it can be concluded that cluster 1 mimics the behavior of legitimate domains while cluster 2 mimics that of DGA domains.
\begin{figure}[!tbp]
	\centering
	\includegraphics[scale=.45,trim=0.65cm 0cm 0cm 1cm]{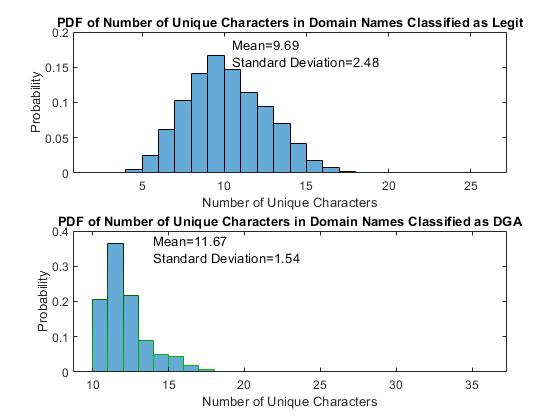}
	\caption{Probability Density Function of Number of Unique Characters For Unlabeled Dataset Using Ensemble Learning Classifier}
	\label{unlabeled_num_of_unique_characters_using_EL}
\end{figure}
\paragraph{Ensemble Learning Classifier}\mbox{}\\
\indent To reduce the number of domains that are identified as possibly suspicious, the information gained from the labeled dataset is applied to the unlabeled dataset. This is done by using the developed ensemble learning classifier to predict the class of each data point in the unlabeled dataset. Fig. \ref{unlabeled_num_of_unique_characters_using_EL} shows the probability density function of the number of unique characters for domains classified as legitimate and DGA respectively. As expected, similar trends observed previously in terms of the mean, standard deviation, and distribution are illustrated. This further emphasizes the validity of these trends. The main difference however is that the number of domains classified as DGA was 29,119 domains. This is almost five times lower than that using the K-means clustering. This is because the information and insights gained from training the ensemble learning classifier on the labeled dataset was used to better identify potentially malicious/suspicious domains. This is based on the fact that K-means grouped the domains into two almost linearly separable clusters despite the fact that they are not as can be deduced from Figs. \ref{labeled_domain_length} and \ref{labeled_num_of_unique_characters}.\\
\indent To further verify the performance of the ensemble classifier, domains are chosen randomly from the subset classified as potentially suspicious and checked using WEBROOT's BrightCloud online tool \cite{url_checker}. This tool gives a reputation score between 0 and 100 to the queried domain based on several factors such as the number of malware infections in the past 12 months, domain popularity, and age. The following three domains are a sample of the domains chosen that had a reputation score below 50 and were classified as suspicious by the BrightCloud tool. 
\begin{itemize}
\item aleximpianti.com
\item a1ukandeuropeancouriers.net
\item aachenhochzeit.de
\end{itemize}
This again shows the merit of the developed ensemble learning classifier as it was indeed able to identify potentially suspicious domains.
\section{Conclusion \& Future Works}\label{conc_dns}
\indent The Domain Name System (DNS) protocol is a core component in todays Internet given that it helps users locate servers, mailing hosts, and other services online \cite{DNS_definition}.
However, due to the lack of data integrity and origin authentication processes within it, it is vulnerable to a variety of security concerns and breaches \cite{DNS_vulnerabilities,DNS_vulnerabilities1}. Among the many vulnerabilities and security challenges of DNS protocol is the issue of typosquatting. Typosquatting refers to the registering of a domain name that is extremely similar to that of an existing popular brand (ex: \textit{www.google.com} and \textit{www.goggle.com}). This is particularly important given that it can be a threat to corporate secrets as well as being used to steal information or commit fraud \cite{DNS_vulnerabilities2}. Therefore, it is crucial that new efficient detection algorithms are developed that can help identify malicious/suspicious queries and protect systems from the various attacks. In this paper, a machine learning-based approach is proposed to tackle the typosquatting DNS vulnerability. Exploratory data analytics were first implemented to better understand the behavior and trends observed in eight domain name-based extracted features. Analysis showed that legitimate domains have a smaller domain name length and a lower number of unique characters when compared to domains generated algorithmically (DGA). Next, a majority voting-based ensemble learning classifier that is based on five traditional supervised machine learning classification algorithms was proposed to identify suspicious domains. Experimental results showed that the developed ensemble learning classifier performed better in terms of accuracy, precision, and F-score while maintaining a high recall value.
Additionally, the observed trends were then verified by studying the same features in an unlabeled dataset using unsupervised machine learning clustering algorithm and through applying the developed ensemble learning classifier. Clustering results showed that the same trends are observable. However, the number of domains identified as potentially suspicious was extremely high (almost half of the dataset). To that end, the information gained from training the classifier on the labeled dataset was leveraged by applying it to the unlabeled dataset. Results showed that when using the developed ensemble learning classifier, the number of domains identified as potentially suspicious was reduced by almost a factor of five while still maintaining the same trends in terms of statistics as seen from the probability density function, mean, and standard deviation.\\
\indent To further build upon this work, several research directions can be followed. The first is considering more features such as query sizes and timing. This can help to identify other types of attacks. Another possibility is to combine several techniques together. For example, time series analysis can be implemented in addition to exploratory data analytics techniques to further improve our understanding of the behavior of the data.

\small
\bibliographystyle{IEEEtran}
\bibliography{Ref1}
\end{document}